\documentclass[a4paper,fleqn]{cas-dc}

\usepackage{graphicx}
\usepackage{amsmath}
\usepackage{amssymb}
\usepackage{booktabs}
\usepackage{amsfonts}
\usepackage{multirow}
\usepackage{times}
\usepackage{epsfig}
\usepackage{graphicx}
\usepackage{amsmath}
\usepackage{amssymb}
\usepackage{adjustbox}
\usepackage{multirow}
\usepackage{multicol}
\usepackage{booktabs}
\usepackage{url}
\usepackage{array}
\usepackage[accsupp]{axessibility}  
\usepackage{hyperref}
\usepackage{pifont}
\usepackage[capitalize]{cleveref}

\usepackage[numbers,sort&compress]{natbib}

\def\tsc#1{\csdef{#1}{\textsc{\lowercase{#1}}\xspace}}
\tsc{WGM}
\tsc{QE}

\newlength\savewidth
\newcommand{\tablestyle}[2]{\setlength{\tabcolsep}{#1}\renewcommand{\arraystretch}{#2}\centering\footnotesize}
\newcommand{\cmark}{\ding{51}}%
\newcommand{\xmark}{\ding{55}}%

\newcommand{\Red}[1]{\textcolor{red}{#1}}
\newcommand{\Blue}[1]{\textcolor{blue}{#1}}

\begin{document}
\let\WriteBookmarks\relax
\def\floatpagepagefraction{1}
\def\textpagefraction{.001}

\shorttitle{}
%
\shortauthors{O.N Manzari et~al.}
\title{MedViT: A Robust Vision Transformer for Generalized Medical Image Classification}

\author[1]{Omid Nejati Manzari}[type=editor,
                        auid=000,bioid=1,
                        ]
\cormark[1]
\ead{omid_nejaty@alumni.iust.ac.ir}
\credit{Conceptualization, Software, Writing- Original draft preparation, Validation, Resources}

\address[1]{School of Electrical Engineering, Iran University of Science and Technology, Tehran, Iran}
\address[2]{Lane Department of Computer Science and Electrical Engineering, West Virginia University, Morgantown, USA}

\author[1]
{Hamid Ahmadabadi}
\credit{Methodology, Writing- Reviewing and Editing}

\author[2]{Hossein Kashiani}

\credit{Writing- Reviewing and Editing, Modification for the final layout}

\author[1]{Shahriar B. Shokouhi} 
\credit{Supervision, Review \& Editing}

\author[1]{Ahmad Ayatollahi}

\credit{Supervision, Review \& Editing}

\cortext[cor1]{Corresponding author.}

\begin{abstract}
Convolutional Neural Networks (CNNs) have advanced existing medical systems for automatic disease diagnosis. However, there are still concerns about the reliability of deep medical diagnosis systems against the potential threats of adversarial attacks since inaccurate diagnosis could lead to disastrous consequences in the safety realm. In this study, we propose a highly robust yet efficient CNN-Transformer hybrid model which is equipped with the locality of CNNs as well as the global connectivity of vision Transformers. To mitigate the high quadratic complexity of the self-attention mechanism while jointly attending to information in various representation subspaces, we construct our attention mechanism by means of an efficient convolution operation. Moreover, to alleviate the fragility of our Transformer model against adversarial attacks, we attempt to learn smoother decision boundaries. To this end, we augment the shape information of an image in the high-level feature space by permuting the feature mean and variance within mini-batches. With less computational complexity, our proposed hybrid model demonstrates its high robustness and generalization ability compared to the state-of-the-art studies on a large-scale collection of standardized MedMNIST-2D datasets.
\end{abstract}




\maketitle

\section{Introduction}

Medical image classification is a critical step in medical image analysis that uses different factors such as clinical information or imaging modalities to differentiate across medical images. A dependable medical image classification may help clinicians evaluate medical images quickly and with less error. The healthcare industry has significantly benefited from recent Convolution Neural Networks (CNNs) advancements. Such advancements have prompted much research into the use of computer-aided diagnostic systems~\cite{lo2022computer, hu2022gashissdb, hu2022application, yang2021computer} based on artificial intelligence in clinical settings. CNNs are able to learn robust discriminative representation from vast volumes of medical data to generate accurate diagnostic performance in medical fields. They validate their satisfactory prediction capabilities and obtain comparable performance as clinicians.

\begin{figure}
    \vspace{12pt}
    \centering
    \includegraphics[width=1.\linewidth]{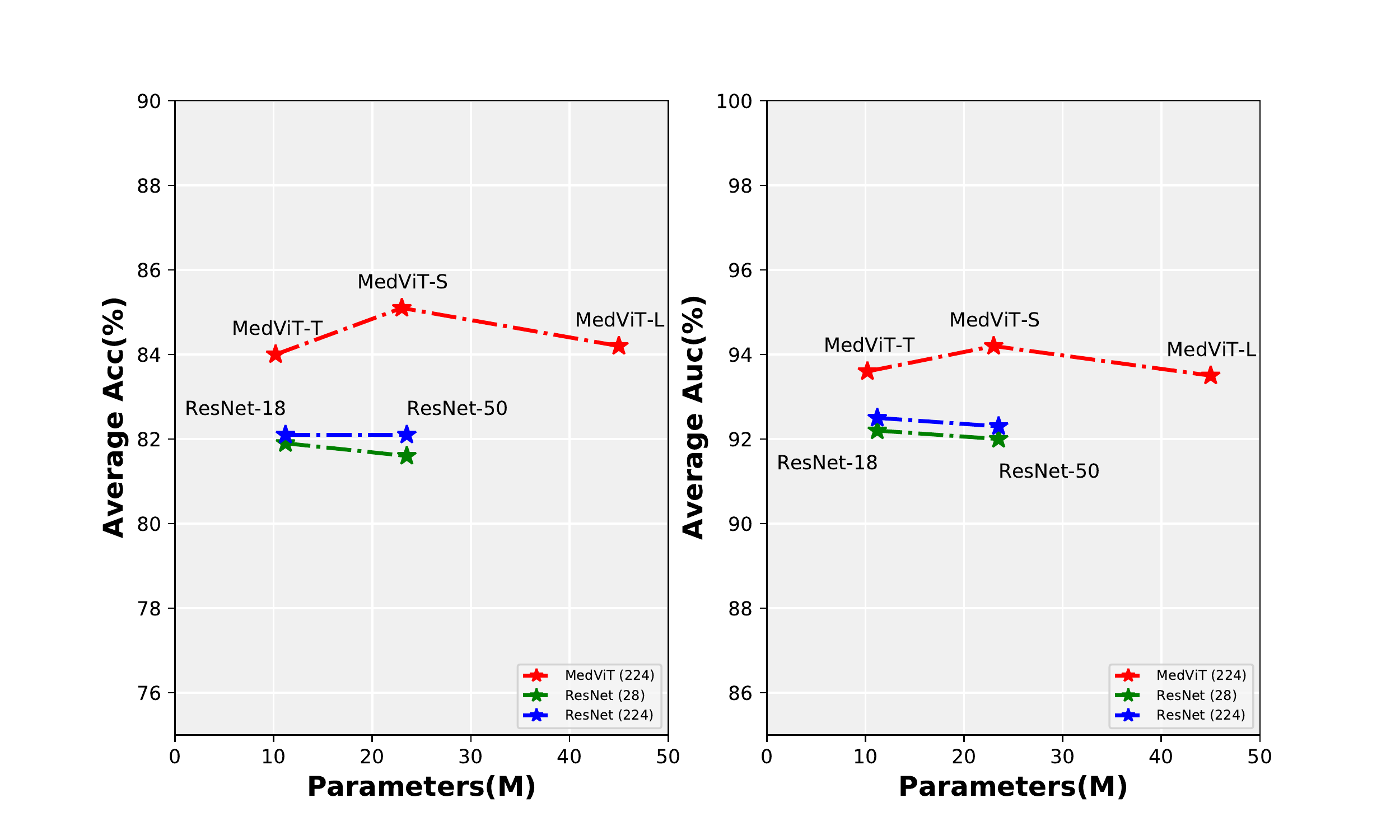}
    \caption{Comparison between MedViTs and the baseline ResNets, in terms of average ACC-Parameters and average AUC-Parametrs trade-off over all 2D datasets.}
    \label{fig:teaser1}
    \vspace{-1em}
\end{figure}

However, the locality bias of CNNs makes it hard for them to learn long-range dependencies in visual data. The texture, shape, and size of many organs vary widely across people, making it difficult to correctly analyze medical data~\cite{igarashi2020anatomical, togo2020deep}. As such, it is important to extract robust feature representation which can model long-range dependencies in different domains for medical image analysis. Recently, the Transformer architectures have adopted the self-attention mechanisms to model the long-range dependencies between input images and have achieved promising results. Different studies demonstrate their performance superiority compared to CNN architectures~\cite{ViT, PVT_v1}. However, a sizable amount of training data is crucial to their success. The construction of a large-scale dataset needs a significant amount of time and resources.

Regarding the medical field, radiologist experts must manually annotate and verify medical data, which is costly and time-consuming. While the Transformer architecture mitigates the shortcomings of CNNs, its computational complexity grows quadratically with spatial or embedding dimensions, therefore making it infeasible for most image restoration tasks involving high-resolution images. That is to say, the Transformer architectures address the long-range dependency modeling in CNNs, yet their computational complexity increases quadratically with the spatial dimension \cite{han2022survey}. As a result, they cannot be used in realistic clinical settings. Moreover, the state-of-the-art studies assume that training and test data are identically distributed. Consequently, on out-of-domain target domains, they often suffer significant performance drops. The domain shift is more pronounced in healthcare areas since medical images can be captured by different devices at various sites. Consequently, due to different scanners and imaging protocols, their data distribution can greatly vary. In addition, variations in epidemiology at different sites could impact the distribution of ground truth labels between various populations \cite{dou2019domain,liu2020ms}.

In this study, we aim to address the above-mentioned challenges and propose a generalized Transformer architecture for medical image analysis. While recent studies in medical image analysis work specifically on the pre-determined medical test sets, our proposed model generalizes to a wide range of medical domains such as CT, X-ray, ultrasound, and OCT domains. To this end, we follow the hierarchical hybrid architecture equipped with a patch embedding layer and a series of convolution and Transformer blocks in each stage with efficient computational complexity. Inspired by recent advances, we cater out Transformer architecture in such a way that each stage consists of two efficient phases to long-term dependencies and model short-term in visual data. In the first phase, we leverage a multi-head convolutional attention block to learn affinity between different tokens in representation subspaces for effective local representation learning. The attention block alleviates the high computational attention-based token mixer in conventional Transformer architectures, thereby improving inference speed.

Compared to the conventional Transformer architectures that suffice to incorporate locality bias into the lower layers of Transformer architectures through tokenization and self-attention components (CvT~\cite{CvT}, Co-Scale Conv-Attentional~\cite{CoaT}, CMT~\cite{Cmt}). We also propose a local feed-forward network (LFFN) that encodes the local dependencies between nearby pixels in feed-forward components of Transformer architectures in all stages. For this objective, a depth-wise convolution is applied to the reshaped 2D feature map. While recent Transformer-based studies have demonstrated a high capacity to learn long-range dependencies between visual data, they fail to encode high-frequency context in visual data. To mitigate this issue, in the second phase of our proposed architecture, we first encode low- and high-frequency feature representation separately with an efficient multi-head self-attention block and a multi-head convolutional attention block, respectively. Then, the computed feature representations are fused and fed to the LFFN to enhance global and local modeling capacity further. As depicted in Figure~\ref{fig:teaser1}, our model shows great superiority in terms of accuracy-complexity against CNNs.

The adversarial attack is a serious security risk for deep neural networks because it could trick the trained models into making incorrect predictions via small, undetectable perturbations. When it comes to healthcare, an adversarial attack could also pose severe security concerns~\cite{ma2022regularization}. The nature of medical data could provide an opportunity for a higher attack success rate with imperceptibility. In this paper, to enhance the adversarial robustness of our proposed Transformer model, we make our model focus more on global structure features (such as shape and style) rather than texture information. Several works~\cite{li2021feature, kim2020puzzle} have found that neural networks tend to rely upon texture information for making predictions, which consequently makes them vulnerable to out-of-distribution samples. Motivated by these studies, we try to encourage our model to rely more on global structure features rather than texture information to boost the generalization performance as well as adversarial robustness. With this objective in mind, we extract the mean and variance of training instances across channel dimensions in feature space and interpolate them with each other. It is worth pointing out that the decision boundary is often sharp, and there is a significant portion of the hidden representation space that is associated with high-confidence predictions~\cite{cao2022survey,verma2019manifold}. With the proposed interpolation, we can explore new useful regions of the feature space, which are mainly relevant to global structure features. This ultimately would enable us to learn smoother decision boundaries, which are beneficial for adversarial robustness and generalization performance.

\section{Related Works}
\label{sec:2}
\textbf{Convolutional networks.} Convolutional neural networks (CNNs) have witnessed extraordinary contributions to the vast fields of computer vision in recent years due to their ability to extract deep discriminative features. ResNet~\cite{ResNet} introduced residual connections to CNN and mitigated the vanishing gradient problem, which ensures the model builds deeper to capture high-level features for image classification. MobileNets~\cite{MobileNet_v1} use pointwise convolutions and depthwise separable convolutions to enhance CNN efficiency. In DenseNet~\cite{DenseNet}, skip connections were used between each two layers, and summation was replaced with concatenation for the dense connections of feature maps. ConvNext~\cite{ConvNext} re-introduces core designs of Vision Transformers and employs $7 \times 7$ depthwise convolutions to design robust CNN architecture, which can achieve comparable results with Transformers. ShuffleNet~\cite{ShuffleNet} performs the channel shuffle operation to fuse separated channel information using group convolution.

\textbf{Vision Transformers.} Since original Transformer architecture achieved remarkable results in natural language processing many attempts have been made to use Transformer architecture to vision tasks like image classification~\cite{ViT}, semantic segmentation~\cite{zheng2021rethinking}, and object detection~\cite{DETR}. In particular, the Vision Transformer (ViT) of Dosovistky et al.~\cite{ViT} shows that pure Transformer-based can also achieve promising result on the image classification task. ViT splits the image into patches (a.k.a., tokens) and applies transformer layers to model the global relation among these patches for classification. T2T- ViT~\cite{T2T} mainly improves tokenization in ViT by delicately generating tokens in a soft split manner, which recursively aggregates neighboring tokens into one token to enrich local structure modeling. Swin Transformer~\cite{Swin} performs self-attention in a local window with the shifted window scheme to alternately model in-window and cross-window connection. PiT~\cite{PiT} follows a similar pyramid structure as CNNs, which produces various feature maps through spatial dimension reduction based on the pooling structure of a convolutional layer. Nowadays, researchers are specifically interested in efficient methods, including pyramidal designs, training strategies, efficient self-attention, etc.

\begin{figure*}
    \centering
    \includegraphics[width=\textwidth]{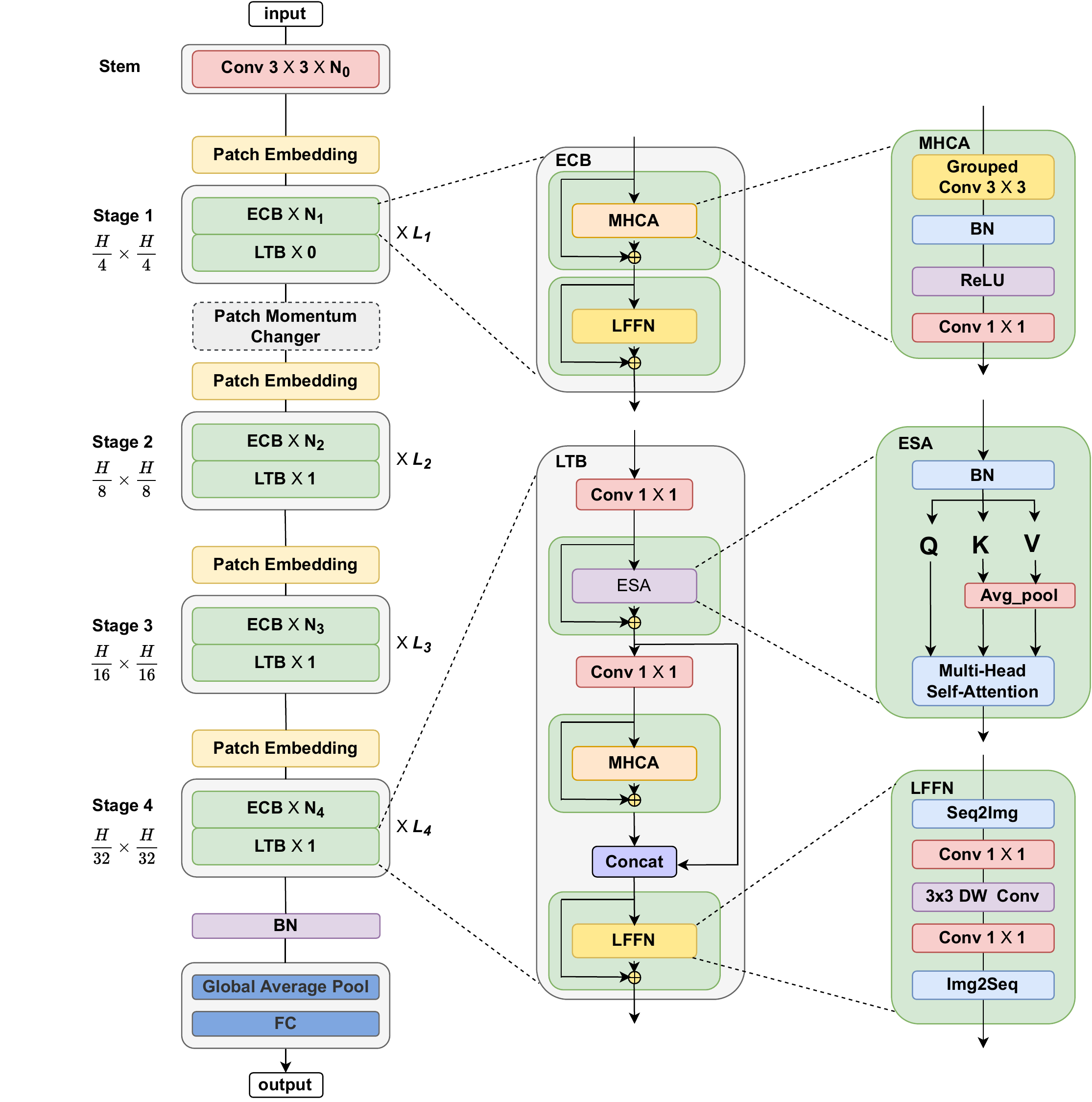}
    \vspace{-4.5mm}
    \caption{\textbf{Overall architecture of the proposed Medical Vision Transformer (MedViT).}}
    \label{fig:overall_arch}
    \vspace{-3.5mm}
\end{figure*}

\textbf{Hybrid Models.} Recent works show that designing a hybrid architecture of transformer and convolution layers helps the model to combine the advantages of both architectures. BoTNet~\cite{BoTNet} uses a slightly-modified self-attention in the last three blocks of ResNet. CMT~\cite{Cmt} block contains depthwise convolution layers based local perception unit and a lightweight transformer block. The CvT \cite{CvT} inserts pointwise and depthwise convolution before self-attention. LeViT~\cite{LeViT} uses the convolutional stem to replace the patch embedding block and achieves fast inference image classification. The MobileViT~\cite{MobileViT} introduces a lightweight vision transformer by combining Transformer blocks with the MobileNetV2~\cite{MobileNet_v2} block in series. Mobile-Former~\cite{Mobile-Former} takes a bidirectional bridge between CNN and transformer to leverage the advantage of global and local concepts.

\begin{figure*}
    \centering
    \includegraphics[width=\textwidth]{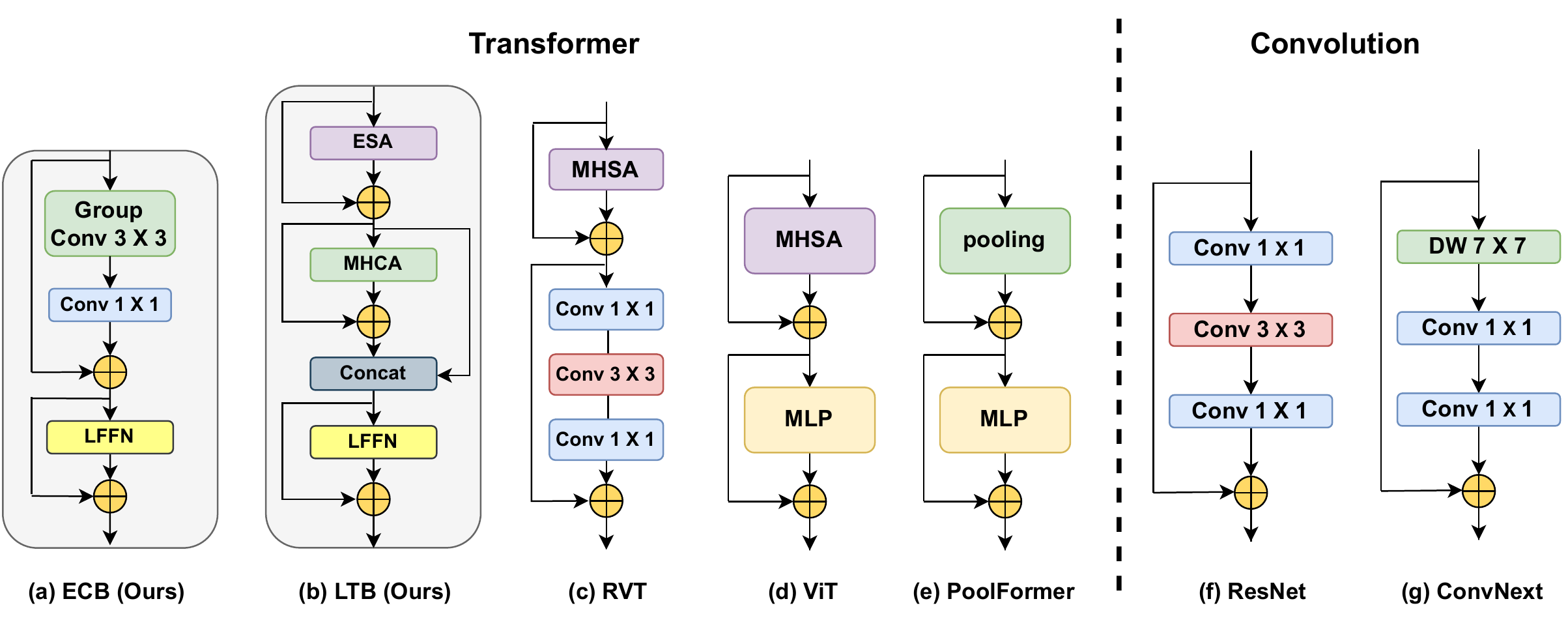}
    \vspace{-5.5mm}
    \caption{{\textbf{Comparison of different core blocks.} Convolution-based that includes the ResNet, ConvNext, and Transformer-based that includes RVT, PoolFormer, ViT, and MedViT (Ours).}}
    \label{fig:Blocks}
    \vspace{-3.5mm}
\end{figure*}

\textbf{Robustness Study.} Due to the nature of convolutional neural networks that rely on low-level features, their assumptions are generally vulnerable to adversarial examples. There are numerous studies on improving the adversarial robustness of CNNs that aim to strengthen it in various approaches. These include carefully designed model~\cite{manzari2023robust, wu2021wider}, strong data augmentation~\cite{rusak2020simple, hendrycks2019augmix}, searched network architecture~\cite{guo2020meet, dong2020adversarially}, improved training strategy~\cite{madry2017towards, li2020shape, xie2020self}, pruning~\cite{ye2019adversarial} of the weights, and quantization~\cite{lin2019defensive}, activation functions~\cite{xie2020smooth} or better pooling~\cite{zhang2019making, vasconcelos2020effective}. Although the methods mentioned earlier perform well on CNNs, there is no evidence that they also improve the adversarial robustness of ViTs.

Following the success of Transformers and their variants in various computer vision tasks, several studies are attempting to examine the robustness of Transformers. Early research conduct the adversarial robustness of Transformers on image classification tasks and compare their vulnerability against the MLP and CNN baselines. The experimental results illustrate that Transformers are more adversarially robust than CNNs~\cite{shao2021adversarial}. Additionally, the adversarial transferability between CNNs and Transformers is unexpectedly low\cite{mahmood2021robustness}. Furthermore, the robustness study~\cite{bhojanapalli2021understanding} of ViTs is extended to the natural distribution shift and common image corruption, which demonstrate the superiority of ViTs over CNNs in the robustness benchmark. Although several studies have challenged the adversarial robustness without carefully designing architecture, in this paper, we do not make a simple comparison of adversarial robustness between CNNs and ViTs, but take a step further by designing a robust hybrid architecture family of MedViTs. Based on the architecture of Transformer, we introduce a novel Augmentation technique to further reduce the fragility of Transformer models.

\section{Method}

We first give a brief overview of the proposed MedViT in this section. Then, We describe the main body designs within MedViT, which include the Efficient Convolution Block (ECB), Local Transformer Block (LTB), and Transformer Augmentation Block (TAB). In addition, we provide different model sizes for the proposed architecture.

\subsection{Overview}

MedViT aims to combine the convolution block and transformer block in a novel approach to achieve a robust hybrid architecture for medical image classification. As shown in Figure~\ref{fig:overall_arch}, MedViT is composed of a patch embedding layer, transformer blocks and a series of stacked convolution in each stage, which follows the hierarchical pyramid architecture traditionally. The spatial resolution will be gradually reduced with a total of $32\times$ ratio by [$4\times, 2\times, 2\times$, and $2\times$] while the channel dimension will be doubled after convolution blocks in each stage. Our purpose in this section is first to explore the core blocks are responsible for embedding multi-scale context and respectively develop robust LTB and ECB to effectively capture long-term and short-term dependencies in input data. LTB also performs the fusion of local and global features, thereby enhancing modeling capabilities. Also we study how to integrate blocks of convolution and transformer technically. Lastly, to further improve the performance and adversarial robustness, we propose a novel Patch Momentum Changer (PMC) data augmentation technique to train our models.

\subsection{Efficient Convolution Block}

We begin by discussing some traditional core blocks of transformer and convolution network, as illustrated in Figure~\ref{fig:Blocks}. To show the effectiveness of the proposed ECB and its superiority over previous methods. ResNet~\cite{ResNet} introduced skip connection and Residual block, which has dominated a wide range of tasks in visual recognition for a long time due to its compatible features and inherent inductive biases for the realistic deployment scenario. Unfortunately, the performance of the Residual block is not satisfactory compared to the Transformer block. The ConvNext block~\cite{ConvNext} constructed from the Residual block by following the designs of the Transformer block without self-attention. Although the ConvNext block enhances network performance to some extent, inefficient components make the model hard to capture high-level structures, such as $7 \times 7$ depth convolution, GELU and LayerNorm. To overcome this, the transformer block has been proposed to capture high-level structures. Transformers have achieved excellent results in various computer vision tasks, and their inherent superiority is jointly conferred by the attention-based token mixing operation~\cite{Swin}. However, these methods merely focus on the model complexity and the standard accuracy. The results demonstrate that the models are vulnerable to adversarial attacks~\cite{bortsova2021adversarial, xu2021towards}, which are intolerable in clinically relevant medical use cases.

In addition, long-range dependencies are crucial for medical images because the background in medical images is generally scattered~\cite{shamshad2022transformers}, thereby learning long-range dependencies between the pixels corresponding to the background can help the network in preventing misclassification~\cite{valanarasu2021medical}. It can be noted that there is still scope for improvement in capturing global context as the shortcoming of prior methods, which do not focus on this aspect for medical image classification tasks.
To address the adversarial robustness and accurate medical classification, we introduce an Efficient Convolution Block (ECB) that achieves outstanding performance as a transformer-based block while retaining the deployment advantage of the Residual block. As illustrated in Figure~\ref{fig:Blocks}~(a), The ECB follows the hybrid architecture, which has been confirmed as necessary for utilizing the multi-scale information. Meanwhile, an effective attention-based token mixing module is equally important. We design a Locally Feed Forward Network (LFFN) as an efficient way of introducing locality into the network with depth-wise convolution and a Multi-Head Convolutional Attention (MHCA) as an effective token mixer. Inspired by Robust Vision Transformer~\cite{mao2022towards} that analyzed the effect of each component of Transformers in the robustness, we build ECB by combining LFFN and MHCA block in the robust paradigm. The proposed ECB can be formulated as follows:

\begin{equation}
    \begin{aligned}
        \tilde{z}^{l} &=\operatorname{MHCA}\left(z^{l-1}\right)+z^{l-1} \\
        z^{l} &=\operatorname{LFFN}\left(\tilde{z}^{l}\right)+\tilde{z}^{l}
        \end{aligned}
\end{equation}

where $z^{l-1}$ denotes the input from the $l-1$ block, $\tilde{z}^{l}$ and $z^l$ are the outputs of MHCA and the $l$ ECB. We will introduce LFNN in detail in the next section.

\subsubsection{Locally Feed-Forward Network}
\begin{figure}
\vspace{-5pt}
    \centering
    \includegraphics[width=1.\linewidth]{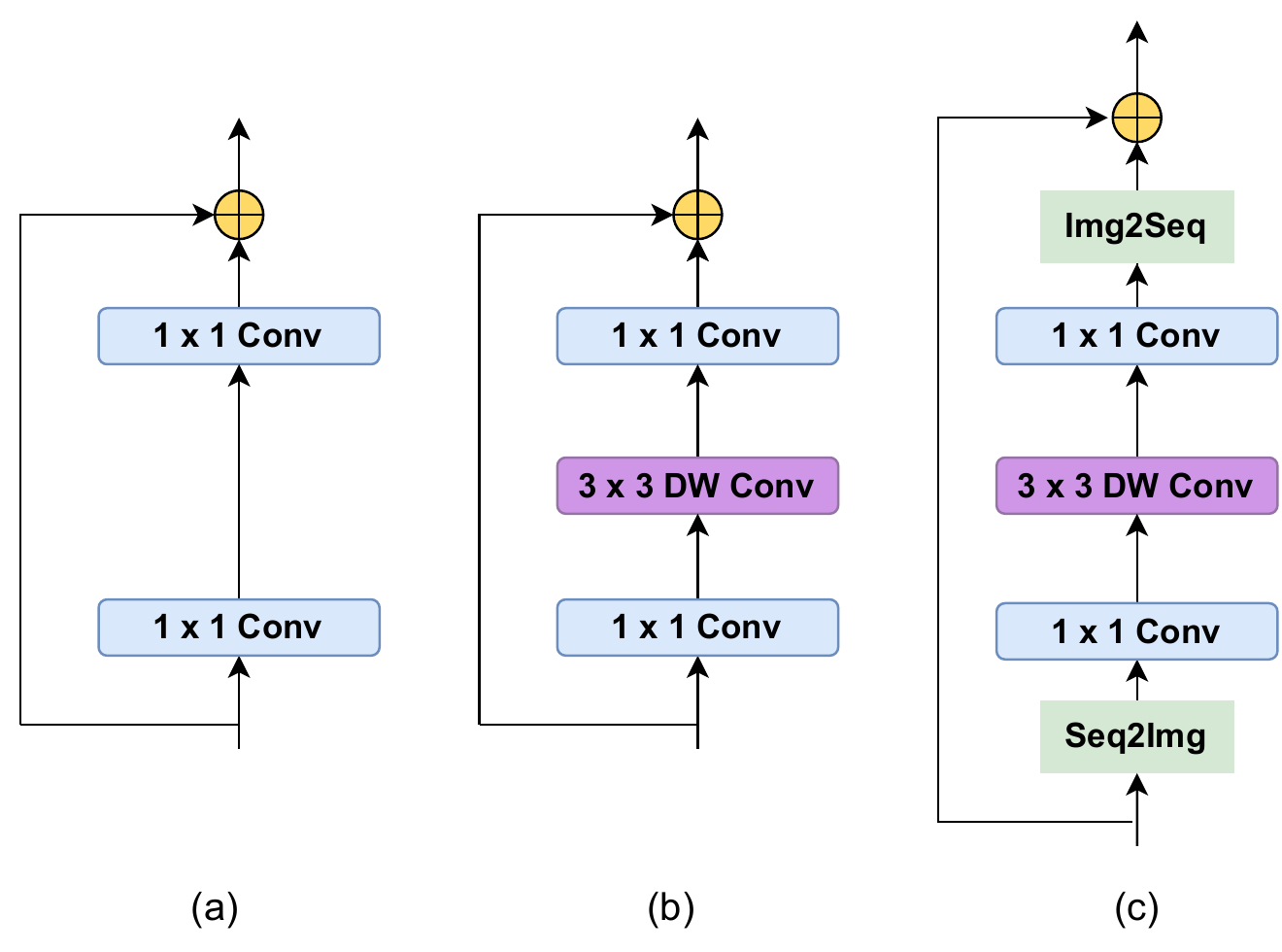}
    \caption{Comparison between the feed-forward network in vision transformers, (a) convolutional feed-forward network. (b) inverted residual block. (c) Our finally utilized network that brings efficient local mechanism into the transformer.}
    \label{fig:LFFN}
    \vspace{-1em}
\end{figure}

The feed-forward network, which is applied position-wise to a sequence of tokens $\mathbf{Z}^{r}$, can be precisely represented by rearranging the sequence of tokens into a 2D lattice, as shown in Figure~\ref{fig:LFFN}~(c). As a result, the reshaped features are represented as follows:

\begin{equation}
\mathbf{Z}^{r}=\operatorname{Seq} 2 \operatorname{Img}(\mathbf{Z}), \mathbf{Z}^{r} \in \mathbb{R}^{h \times w \times d}
\end{equation}

where $h=H/p$ and $w=W/p$. $Seq2IMG$ takes a sequence and converts it into a feature map that can be visualized. The tokens are placed at pixel locations on the feature map, and each token corresponds to one pixel. Through this perspective, it is possible to introduce locality into the network by recovering the proximity between tokens. The fully-connected layers could be replaced by $1 \times 1$ convolution layers, i.e.
\begin{equation}
\begin{aligned}
\mathbf{Y}^{r} &=f\left(\mathbf{Z}^{r} \circledast \mathbf{W}_{1}^{r}\right) \circledast \mathbf{W}_{2}^{r} \\
\mathbf{Y} &=\operatorname{Img} 2 \mathrm{Seq}\left(\mathbf{Y}^{r}\right)
\end{aligned}
\end{equation}

where $\mathbf{W}_{1}^{r} \in \mathbb{R}^{d \times \gamma d \times 1 \times 1}$ and $\mathbf{W}_{2}^{r} \in \mathbb{R}^{{\gamma} d \times d \times 1 \times 1}$ are reshaped from $W_1$ and $W_2$ and denote the kernels of convolutional layers. With $Img2Seq$, the image feature map is converted back into a token sequence, which is then used by the next self-attention layer by transforming it into the fused token.

\subsection{Local Transformer Block}
\label{sec:2.2}
While the local representation has been effectively learned through the ECB, capturing global information is urgent and needs to be addressed in this block. It is well known that transformer blocks have the capability to capture low-frequency signals, which are very useful for capturing global information (e.g., global shapes and structures). However, There have been a few related studies~\cite{park2022vision} that have demonstrated that transformer blocks have a tendency to deteriorate high-frequency information, such as information about the local texture of objects, to some extent. It is essential that signals in different frequency segments are fused in order to extract essential and distinct features in the computer vision system~\cite{kauffmann2014neural}.

In response to these observations, we have developed the Local Transformer Block (LTB) in order to capture multi-frequency signals in a lightweight mechanism with high efficiency. Moreover, LTB works as an effective multi-frequency signal mixer, thus enhancing the overall modeling capability of the network. As shown in Figure~\ref{fig:Blocks}~(b), LTB first captures low-frequency signals by utilizing an Efficient Self Attention (ESA), which can be formulated as follows:
\begin{equation}
\begin{aligned}
\operatorname{ESA}(x) &=\operatorname{Concat}\left(\operatorname{SA}_{1}\left(x_{1}\right), \mathrm{SA}_{2}\left(x_{2}\right), \ldots, \mathrm{SA}_{h}\left(x_{h}\right)\right) W^{O} \\
\operatorname{SA}(X) &=\operatorname{Attention}\left(X \cdot W^{Q}, \mathrm{P}_{s}\left(X \cdot W^{K}\right), \mathrm{P}_{s}\left(X \cdot W^{V}\right)\right)
\end{aligned}
\end{equation}

where $X = [x_1, x_2, ..., x_h]$ denotes to divide the input feature X into multi-head form in channel dimension. $ W^O $ is an output projection layer and $ h $ is the number of heads. In order to reduce the spatial resolution of self-attention, SA was derived from linear SRA~\cite{PVT_v2}.
Attention calculates $W^Q, W^K, W^V$  linear layers in standard attention form as $Attention(Q,K,V) = softmax(QK^T / \sqrt{d})V$, in which $d$ is the transformer hidden dimension. The $Ps$ operation involves an avg-pool operation with a stride $s$ parameter for downsampling the spatial dimensions before the attention operation is applied to reduce the computation cost. Furthermore, we observe that the number of channels in the ESA module is also a major determinant of the time consumption of the module. With the help of point-wise convolutions, LTB enables further acceleration of inference by reducing the dimension of the channel before it passes to the ESA module. In order to reduce the number of channels, a shrinking ratio r is introduced. Additionally, the ESA module also utilizes Batch Normalization to make the module's deployment extremely efficient.

\begin{table}[]
  \centering
  \caption{Detailed configurations of MedViT variants. C and S denotes number of channels and stride of convolution for each stage.}
  \label{tab:configs}
  \resizebox{0.45\textwidth}{!}{
  \begin{tabular}{ccc|c|c|c}
  \toprule
  Stages                        & Output size                                         & Layers                                                          & MedViT-T                & MedViT-S            & MedViT-L        \\  \midrule
  \multirow{5}{*}{Stem}         & \multirow{5}{*}{$\displaystyle{\frac{H}{4}} \times \frac{W}{4} $}  & \multirow{5}{*}{\shortstack{Convolution \\ Layers}}             & \multicolumn{3}{c}{$\text{Conv}\ 3\times3, C=64, S=2$}                       \\  \cmidrule(l){4-6}
                                &                                                     &                                                                 & \multicolumn{3}{c}{$\text{Conv}\ 3\times3, C=32, S=1$}                       \\  \cmidrule(l){4-6}
                                &                                                     &                                                                 & \multicolumn{3}{c}{$\text{Conv}\ 3\times3, C=64, S=1$}                       \\  \cmidrule(l){4-6}
                                &                                                     &                                                                 & \multicolumn{3}{c}{$\text{Conv}\ 3\times3, C=64, S=2$}                       \\  \midrule
  \multirow{4}{*}{Stage 1}      & \multirow{4}{*}{$\displaystyle{\frac{H}{4}} \times \frac{W}{4} $}  & \multirow{1}{*}{Patch Embedding}                                & \multicolumn{3}{c}{$\text{Conv}\ 1\times1, C=96$}                            \\  \cmidrule(l){3-6}
                                &                                                     & \multirow{2}{*}{\shortstack{MedViT \\ Block}}            & \multicolumn{3}{c}{\multirow{2}{*}{$\begin{bmatrix} \text{ECB} \times 3, 96 \end{bmatrix} \times 1$}}                    \\
                                &                                                     &                                                                 & \multicolumn{3}{c}{}                                                                                          \\  \midrule
  \multirow{6}{*}{Stage 2}      & \multirow{6}{*}{$\displaystyle{\frac{H}{8}} \times \frac{W}{8} $}  & \multirow{2.5}{*}{Patch Embedding}                              & \multicolumn{3}{c}{$\text{Avg\_pool}, S=2$}                                                                          \\  \cmidrule(l){4-6}
                                &                                                     &                                                                 & \multicolumn{3}{c}{$\text{Conv}\ 1\times1, C=192$}                                                                   \\  \cmidrule(l){3-6}
                                &                                                     & \multirow{3}{*}{\shortstack{MedViT \\ Block}}            & \multicolumn{3}{c}{\multirow{3}{*}{$\begin{bmatrix} \text{ECB} \times 3, 192 \\ \text{LTB} \times 1, 256 \end{bmatrix} \times 1$}}    \\
                                &                                                     &                                                                 & \multicolumn{3}{c}{}                                                                                          \\
                                &                                                     &                                                                 & \multicolumn{3}{c}{}                                                                                          \\  \midrule
  \multirow{6}{*}{Stage 3}      & \multirow{6}{*}{$\displaystyle{\frac{H}{16}} \times \frac{W}{16}$} & \multirow{2.5}{*}{Patch Embedding}                              & \multicolumn{3}{c}{$\text{Avg\_pool}, S=2$}                                                                          \\  \cmidrule(l){4-6}
                                &                                                     &                                                                 & \multicolumn{3}{c}{$\text{Conv}\ 1\times1, C=384$}                                                                   \\  \cmidrule(l){3-6}
                                &                                                     & \multirow{3}{*}{\shortstack{MedViT \\ Block}}            & \multirow{3}{*}{$\begin{bmatrix} \text{ECB} \times 4, 384 \\ \text{LTB} \times 1, 512 \end{bmatrix} \times 2$}        & \multirow{3}{*}{$\begin{bmatrix} \text{ECB} \times 4, 384 \\ \text{LTB} \times 1, 512 \end{bmatrix} \times 4$}        & \multirow{3}{*}{$\begin{bmatrix} \text{ECB} \times 4, 384 \\ \text{LTB} \times 1, 512 \end{bmatrix} \times 6$}       \\
                                &                                                     &                                                                 &                                                                                               &                                                                                               &                                                                                              \\
                                &                                                     &                                                                 &                                                                                               &                                                                                               &                                                                                              \\  \midrule
  \multirow{6}{*}{Stage 4}      & \multirow{6}{*}{$\displaystyle{\frac{H}{32}} \times \frac{W}{32}$} & \multirow{2.5}{*}{Patch Embedding}                              & \multicolumn{3}{c}{$\text{Avg\_pool}, S=2$}                                                                          \\  \cmidrule(l){4-6}
                                &                                                     &                                                                 & \multicolumn{3}{c}{$\text{Conv}\ 1\times1, C=768$}                                                                   \\  \cmidrule(l){3-6}
                                &                                                     & \multirow{3}{*}{\shortstack{MedViT \\ Block}}            & \multicolumn{3}{c}{\multirow{3}{*}{$\begin{bmatrix} \text{ECB} \times 2,  \ 768 \\ \text{LTB} \times 1, 1024 \end{bmatrix} \times 1$}}    \\
                                &                                                     &                                                                 & \multicolumn{3}{c}{}                                                                                          \\
                                &                                                     &                                                                 & \multicolumn{3}{c}{}                                                                                          \\  \bottomrule
  \end{tabular}
}
\vspace{-0.25cm}
\end{table}

It is significant to note that LTB has a multi-frequency configuration that is designed to function in conjunction with ESA and MHCA modules. Following that, we design a new attention mechanism that is based on efficient convolutional operations for improving the efficiency of the LTB. Inspired by the effective multi-head design in MHSA~\cite{wu2021wider}, we build our convolutional attention (CA) with multi-head paradigm, which jointly attends to information from different representation subspaces at different positions for effective local representation learning. The proposed MHCA can be formulated as follows:
\begin{equation}
\begin{aligned}
\operatorname{MHCA}(x) &=\operatorname{Concat}\left(\mathrm{CA}_1\left(x_1\right), \mathrm{CA}_2\left(x_2\right), \ldots, \mathrm{CA}_h\left(x_h\right)\right) W^O \\
\mathrm{CA}(X) &= \left(\mathrm{W} \cdot \mathrm{T}_{\{i,j\}}\right), \text { where } \mathrm{T}_{\{i,j\}} \in X
\end{aligned}
\end{equation}

where MHCA captures information from h parallel representation subspaces and CA is single-head convolutional attention. $W$ is trainable parameter and $\mathrm{T}_{\{i,j\}}$ are adjacent tokens in input feature $X$. The CA is calculated by the inner product operation of displaced vectors between adjacent tokens $\mathrm{T}_{\{i,j\}}$ and trainable parameter W. Since the multi-head self-attention (MHSA) in Transformers could capture the global context, we propose the CA from the MHCA, which can learn affinity between different tokens in the local receptive. Notably, our implementation of MHCA involves a point-wise convolution and a group convolution (multi-head convolution), as shown in Figure~\ref{fig:Blocks}~(a).

the output features of the MHCA and the ESA are concatenated to produce a mix of high-low features. As a final step, an MLP layer is borrowed at the end of the process in order to extract the essential and distinct features. In brief, the implementation of the LTB can be summarized as follows:
\begin{equation}
\begin{aligned}
\bar{z}^{l} &=\operatorname{Proj}\left(z^{l-1}\right) \\
\ddot{z}^{l} &=\mathrm{ESA}\left(\bar{z}^{l}\right)+\bar{z}^{l} \\
\dot{z}^{l} &=\operatorname{Proj}\left(\ddot{z}^{l}\right) \\
\tilde{z}^{l} &=\operatorname{MHCA}\left(\dot{z}^{l}\right)+\dot{z}^{l} \\
\hat{z}^{l} &=\operatorname{Concat}\left(\ddot{z}^{l}, \tilde{z}^{l}\right) \\
z^{l} &=\operatorname{LFFN}\left(\hat{z}^{l}\right)+\hat{z}^{l}
\end{aligned}
\end{equation}

where ${z}^{l}$ is the output of LTB from the $l$-th block, and $\tilde{z}^{l}$, $\ddot{z}^{l}$ denote the output of MHCA and ESA, respectively. Proj refers to the point-wise convolution layer associated with the project channel. In order to provide efficient norm and activation layers for LTB, the BN and the ReLU are uniformly adopted instead of the LN and the GELU as the efficient norm and activation layers. A major advantage of the LTB over traditional transformer blocks is its ability to capture and mix multi-frequency information in such a lightweight mechanism, so the performance of the model is greatly enhanced.


\begin{figure*}
    \centering
    \includegraphics[width=\textwidth]{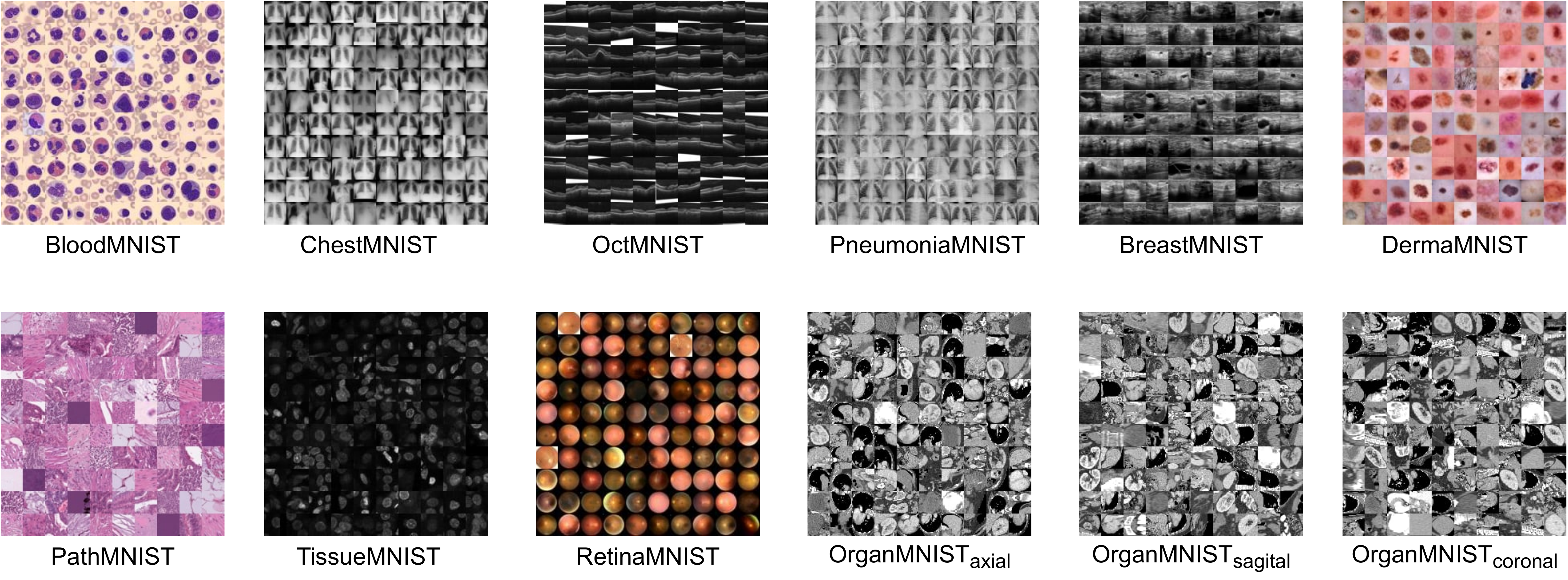}
    \vspace{-4.5mm}
    \caption{\textbf{MedMNIST-2D Classification}. MedMNIST is a collection of 12 pre-processed medical image datasets. It is designed to be educational, standardized, diverse and lightweight, which could be used as a general classification benchmark in medical image analysis.}
    \label{fig:datasets}
    \vspace{-3.5mm}
\end{figure*}

\subsection{Transformer Augmentation Block}
Image augmentation techniques apply geometric transformation functions such as rotating, cropping, and flipping or color space transformation functions such as edge enhancement, grayscale transformations, and color jittering on an input image. Data augmentation is an important strategy for ViTs because they suffer from data scarcity when trained on relatively small-size datasets, while is a data-space solution to the problem of limited data can be solved by strong data augmentation~\cite{chen2021amplitude}. Moreover, a rich data augmentation also helps with robustness and generalization, which has been verified in previous works~\cite{carratino2020mixup, chen2020group, kim2020puzzle}. In order to improve the diversity of the augmented training data, we introduce Patch Momentum Changer (PMC) augmentation for ViTs, which blends feature normalization with data augmentation at training time for a pair of images at token level. Our motivation stems from the fact that all layers of ViTs  have global receptive fields, so they are also concerned with the local relationships that exist between the tokens. We believe that the traditional augmentations, which randomly transform the whole image in order to enlarge the data, are sufficient for providing global context. However, for ViTs that naturally capture global receptive fields, conventional augmentations are less beneficial. In order to increase the interactions between the tokens, PMC laterally fuses intermediate feature maps and targets across two training samples. It mixes two very different components, the feature moments of one instance are combined with the normalized features of another at token level. This asymmetric composition in feature space helps Transformer to improve robustness and generalization when predicting medical image datasets.

At each stage, After feeding the word-level features $\hat{Z}^{l}$ into the Locally Feed-Forward Network, PMC  could take as input the feature representation $Z^l$, which is a 2D tensor. Similar to Cutmix and Mixup, features of two random training samples are fused with their labels, while performing the feature normalization. Specifically, PMC combine the normalized feature map of one sample with the feature moments of another. This nonsymmetric combination in word-level feature aims to create robust targets and smooth out the decision boundary of the trained classifier. To normalize features at different stages inside the MedViT model, function $F$ is defined. This function takes the word-level features $Z_i^l$ of the $i-th$ input $x_i$ at stage $l$ of MedViT model and generates three outputs which include: the first-moment $\mu _i$, the second moment $\sigma _i$, and the normalized word-level features $|Z_i^l|$, as follows:

\begin{align}
  && F(Z_i^l) = (\mu_i^l , \sigma_i^l, |Z_i^l| ).
\end{align}

Function $F$ calculates the value of the first and second momentum after feeding the word-level feature $Z_i^l$ through the LFFN module. This operation relatively resembles PONO function~\cite{li2021feature} in the realm of CNNs. To employ MedViT model, we randomly select two different images $x_A$ and $x_B$. The operation could apply at each stage of the model, but it is more effective at the first stage. Consequently, we drop the $l$ superscript for notational simplicity. Augmented features are generated from normalized word-level features of the first image $(x_A : F(Z_A) = (\mu _A , \sigma _A , |{Z}_A| ))$ that are combined with the moments of the second image $(x_B : F(Z_B) = (\mu _B , \sigma _B , |{Z}_B|))$  as follows:

\begin{align}
  && Z_A^{(B)} = \sigma _B \frac{|{Z}_A| - \mu _A}{\sigma _A} + \mu _B,
\end{align}
where $Z_A^{(B)}$ are augmented features and  $ |{Z}_A|,\mu _A,\sigma _A $ are the normalized word-level features, the first-moment, and the second moment of image $A$. In addition, $ \mu _B$ and $\sigma _B $ are the first and second moments of image $B$. The model continues the forward pass from stage $l$ until the output using these features $Z_A^{(B)}$. The lost function is modified to force the model to pay attention to injected features of image $x_B$. The mixed new loss function would be created as follows:

\begin{align}
  && \lambda \cdot  (Z_A^{(B)}, y_A) + (1- \lambda) \cdot (Z_A^{(B)}, y_B),
\end{align}
where $ \lambda \in (0,1) $ is a fixed variable for setting the combination of the features and the moments.  Also $(y_A,y_B)$ are labels of images, which are combined together for the final loss.

PMC is performed entirely at the feature level inside the transformer vision network and can be readily combined with other augmentation methods that operate on the raw input (pixels or words). We explicitly encourage the transformer to encode better long-range dependency to correctly classify the image with the feature of another image combined inside. We show that our approach can lead to consistent accuracy gain when used in MedViT, and also enhances the adversarial robustness of the transformers. Besides, we evaluate the efficacy of PMC thoroughly across several datasets.

\begin{table*}[!ht]
\scriptsize
\centering
\caption{{\bf Overview of MedMNIST v2~\cite{yang2021medmnist} dataset}. MedMNIST2D consists of 12 biomedical datasets of 2D images. Some of the notations used in datasets include OR: Ordinal Regression. MC: Multi-Class. ML: Multi-Label. BC: Binary-Class.}
\vspace{-0.7mm}
\label{tab:overview}
\resizebox{\textwidth}{!}{
\begin{tabular}{lcccccc}
\toprule
Name  & Data Modality & Task (\# Classes / Labels) & \# Samples & \# Training / Validation / Test \\
\midrule
\textbf{MedMNIST2D}\\
ChestMNIST & Chest X-Ray & ML (14) BC (2) & 112,120 & 78,468 / 11,219 / 22,433   \\
PathMNIST &  Colon Pathology & MC (9) & 107,180 & 89,996 / 10,004 / 7,180 \\
OCTMNIST & Retinal OCT & MC (4) & 109,309 & 97,477 / 10,832 / 1,000  \\
DermaMNIST & Dermatoscope & MC (7) & 10,015 & 7,007 / 1,003 / 2,005 \\
RetinaMNIST & Fundus Camera & OR (5) & 1,600 & 1,080 / 120 / 400 \\
PneumoniaMNIST & Chest X-Ray & BC (2) & 5,856 &  4,708 / 524 / 624 \\
BreastMNIST & Breast Ultrasound & BC (2) & 780 & 546 / 78 / 156  \\
TissueMNIST & Kidney Cortex Microscope & MC (8) & 236,386 & 165,466 / 23,640 / 47,280  \\
BloodMNIST & Blood Cell Microscope & MC (8) & 17,092 & 11,959 / 1,712 / 3,421  \\
OrganAMNIST & Abdominal CT & MC (11) & 58,850 & 34,581 / 6,491 / 17,778  \\
OrganCMNIST & Abdominal CT & MC (11) & 23,660 & 13,000 / 2,392 / 8,268  \\
OrganSMNIST & Abdominal CT & MC (11) & 25,221 & 13,940 / 2,452 / 8,829 \\
\bottomrule
\end{tabular}
}
\end{table*}

\section{Experiments}

\subsection{Datasets}
MedMNIST datasets include a set of 12 pre-processed datasets that include CT, X-ray, ultrasound and OCT images. These datasets are used in various classification tasks including multi-label, ordinal, multi-class, regression, and binary. The size of the data in this collection varies from at least 100 to more than 100,000. As shown in Table~\ref{tab:overview}, the diversity of these datasets has created a favorable criterion for classification tasks. The pre-processing and split of the datasets into training, validation and test subsets have been done according to~\cite{yang2021medmnist}.

\textbf{PathMNIST} is adapted from a dataset based on kather's work~\cite{kather2019predicting}. This dataset contains 100,000 image patches that are manually divided into 9 different classes. It is also adapted from another dataset that contains 7180 non-overlapping image patches in the classes of fat loss, background, debris, lymphocytes, mucosa, smooth muscle, normal colon mucosa, cancer-related stroma, and epithelium.

\textbf{ChestMNIST} is adapted from a dataset consisting of chest x-ray images~\cite{wang2017chestx}. The dataset consists of 112,120 frontal X-ray images from a total of 32,717 patients. This dataset contains 14 different classes of diseases, which is a multi-class dataset in the MEDMNIST collection. We used benchmark standards to split and resize data.

\textbf{DermaMNIST} is based on HAM10000~\cite{tschandl2018ham10000}, a large
collection of multi-source dermatoscopic images of common pigmented skin lesions. The dataset consists of 10015 dermatoscopic images of a size of 450 × 600. It consists of 7 diagnostic categories as follows: Melanocytic Nevi (NV), Melanoma (MEL),  Basal Cell Carcinoma (BCC), and Intra-Epithelial Carcinoma (AKIEC), Actinic Keratosis, Benign Keratosis (BKL), Vascular lesions (VASC), Dermatofibroma (DF). All formulated as a multi-class classification task.

\textbf{OCTMNIST} is built on the back of a prior set~\cite{dataset20202nd} of 109309 valid optical coherence tomography (OCT) images that were collected for retinal diseases. 4 types are involved, leading to a multi-class classification task.

\textbf{PneumoniaMNIST} is adapted from a prior dataset~\cite{qi2021elastic}. This dataset consists of 5856 pediatric chest X-ray images on just two classes. The task is to categorize pneumonia into two binary classes, pneumonia and normal.

\textbf{RetinaMNIST} is based on DeepDRiD (Deep Diabetic Retinopathy)~\cite{chen2021alleviating}, the dataset provides 628 patients data including 1600 retina fundus images. The task is ordinal regression for 5-level grading of diabetic retinopathy severity.

\textbf{TissueMNIST} is adapted from the Broad Bioimage Benchmark Collection~\cite{ ljosa2012annotated}. This dataset is categorized into 8 classes of human kidney cortex cells, which contains 236,386 segmented images from different reference tissue specimens.

\textbf{BloodMNIST} is adapted from a prior blood collection~\cite{acevedo2020dataset}. The dataset is categorized into 8 classes and contains a total of 17,092 normal blood cell images.

\textbf{BreastMNIST} is based on a dataset~\cite{al2020dataset} of 780 breast
ultrasound images. It is categorized into 3 classes: benign, malignant, and normal. Because low-resolution images are used, the task are simplified into binary classification by combing normal and benign as positive, and classify them against malignant as negative.

\textbf{OrganMNIST} \{Axial, Coronal, Sagittal\} is taken from 3D computed tomography (CT) images by the Liver Tumor Segmentation Benchmark (LiTS)~\cite{acevedo2020dataset}. As a way to get the organ labels, bounding-box annotations of 11 body organs from another study have been used~\cite{ljosa2012annotated}. As a result of translating the Hounsfield-Unit (HU) of the 3D images to greyscale and with an abdominal window, it is then possible to produce 2D images by selecting slices in the axial, coronal, and sagittal directions within the bounding boxes of the 3D images. The view is the only difference between OrganMNIST and LiST. The images are resized into 1 × 28 × 28 to function as targets for multiclass classification of 11 organs of the body. The training and validation sets are comprised of 115 and 16 CT scans from the source training set, respectively. The 70 CT scans from the source test set are also considered the test set.

\begin{figure*}
    \centering
    \includegraphics[width=\textwidth]{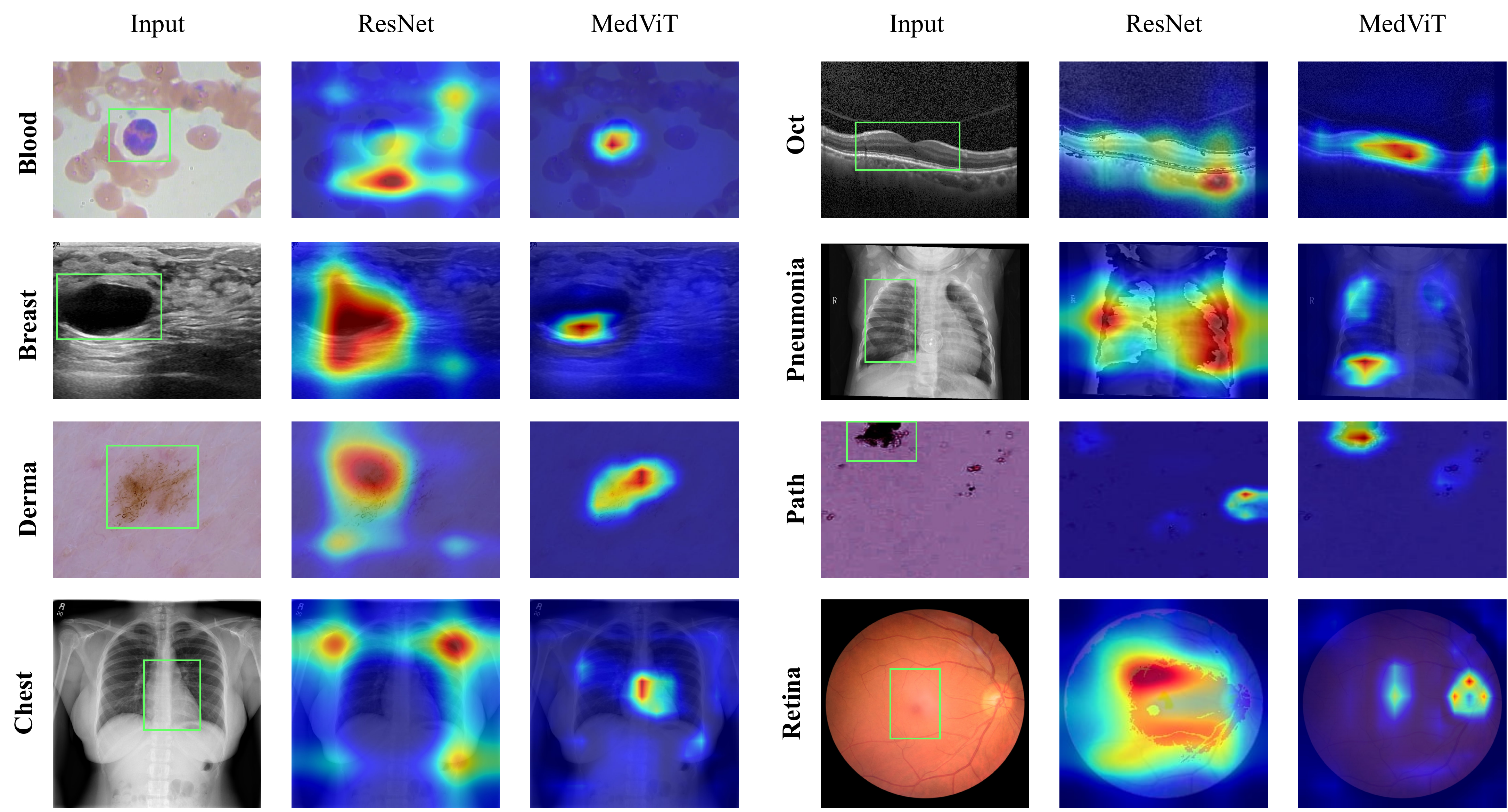}
    \vspace{-4.5mm}
    \caption{Visual inspection of MedViT-T and ResNet-18 using Grad-CAM on MedMNIST-2D datasets. The green rectangles are used to show a specific part of the image that contains information relevant to the diagnosis or analysis of a medical condition, where the superiority of our proposed method can be clearly seen.}
    \label{fig:heatmap}
    \vspace{-3.5mm}
\end{figure*}

\subsection{Implementation Details}
\label{sec:2.5}
Our experiment on medical image classification is conducted on the MedMNIST dataset, which is composed of 12 standardized datasets from comprehensively medical resources covering a range of primary data modalities representative of medical images. To make a fair and objective judgment, we follow the same training settings of the MedMNISTv2~\cite{yang2021medmnist} without making any changes from the original settings. Specifically, we train all of the MedViT variants for 100 epochs on NVIDIA 2080Ti GPUs, and use a batch size of 128. The images are first resized to a size of 224 x 224 pixels. We employ an AdamW optimizer~\cite{AdamW} with an initial learning rate of 0.001, the learning rate is decayed by a factor set of 0.1 in 50 and 75 epochs. Moreover, we introduce MedViT models at three different network sizes MedViT-T, MedViT-S, and MedViT-L, as shown in Table~\ref{tab:configs}. All of them adopt the best settings investigated in section 2 and are trained for each dataset separately. For MedViT$^*$, we add augmentation in the training phase. The PMC uses the mixture feature normalization with data augmentation at training time for each input image patch.

\subsection{Evaluation metric}
We report Accuracy (ACC) and Area under the ROC Curve (AUC) as the standard evaluation metrics. In contrast to AUC, which is a free threshold metric used to evaluate continuous prediction scores, ACC uses a threshold-based metric to evaluate discrete prediction labels. Therefore, ACC is more sensitive to class discrepancy than AUC. Because our experiments have many datasets of different sizes and data variety, both ACC and AUC could serve as comprehensive metrics. Although there are many other metrics, to establish a fair comparison, we select ACC and AUC for the benchmarking methods reported in the original publications~\cite{yang2021medmnist, yang2021med}. We report the results of ACC and AUC for each dataset in table~\ref{tab:2DResults}. Similar to~\cite{yang2021medmnist}, we average the results over MedMNIST2D and report the average AUC and ACC scores in table~\ref{tab:AVG_2DResults}.

\section{Evaluation Results}
\subsection{Results on Each Dataset}
The comparison of the proposal method with previous state-of-the-art (SOTA) methods in terms of the AUC and ACC on each dataset of MedMNIST-2D is shown in Table~\ref{tab:2DResults}. MedViT outperforms previous SOTA methods by a large margin. Compared to AutoML methods, our MedViT-S shows superior learning ability on both evaluation metrics, observing an increase of 2.3\% (AUC) and 3.0\% (ACC) in RetinaMNIST and an increase of 1.1\% (AUC) and 2.8\% (ACC) in TissueMNIST compared to Google AutoML Vision and AutoKeras, respectively. Concretely, MedViT steadily improve the performance of visual classification tasks in the MedMNIST-2D benchmark, particularly for PathMNIST, ChestMNIST, DermaMNIST, PneumoniaMNIST and BreastMNIST.

Although a specific architecture designed for a special image format is more accurate in one area, we have designed MedViT by combining efficient blocks to extract local and global features for generalized medical image classification. Also, MedMNIST-2D contains different types of images, including CT, ultrasound, X-ray, and OCT, which are colour or grayscale with different content in the medical domain. Results in Table~\ref{tab:2DResults} show that our MedViT performs the classification of medical images well for MedMNIST datasets. Besides, the efficiency in terms of the number of parameters is indicated in Table~\ref{tab:AVG_2DResults}, which will be discussed in the following sections. These results demonstrate that the proposed MedViTs design has effectiveness and good generalization ability.

\begin{table*}[!htb]
\footnotesize
	\caption{{\bf Comparison results of the proposed method on the MedMNIST2D} in metrics of AUC and ACC. White background shows CNN-based and AutoML methods, while the proposed MedViT are colored in blue. Also \textcolor{blue}{blue} indicates the best result, and \textcolor{red}{red} displays the second-best.}
	\label{tab:2DResults}
	\begin{center}
		
		\begin{tabular}{l||cc||cc||cc||cc||cc||cc}
			\toprule
			\multirow{2}{*}{\textbf{Methods}} &
			\multicolumn{2}{c||}{\textbf{PathMNIST}} &
			\multicolumn{2}{c||}{\textbf{ChestMNIST}} &
			\multicolumn{2}{c||}{\textbf{DermaMNIST}} &
			\multicolumn{2}{c||}{\textbf{OCTMNIST}} &
			\multicolumn{2}{c||}{\textbf{PneumoniaMNIST}} &
			\multicolumn{2}{c}{\textbf{RetinaMNIST}} \\ \cline{2-13}
			& AUC & ACC & AUC & ACC & AUC & ACC & AUC & ACC & AUC & ACC & AUC & ACC\\ \midrule
			ResNet-18 (28)~\cite{ResNet}     & 0.983 & 0.907 & 0.768 & 0.947 & 0.917 & 0.735 & 0.943 & 0.743 & 0.944 & 0.854 & 0.717 & 0.524 \\
			ResNet-18 (224)~\cite{ResNet}        & 0.989 & 0.909 & 0.773 & 0.947 & \textcolor{red}{0.920} & 0.754 & 0.958 & 0.763 & 0.956 & 0.864 & 0.710 & 0.493 \\
			ResNet-50 (28)~\cite{ResNet}         & 0.990 &  0.911 & 0.769 & 0.947 & 0.913 & 0.735 & 0.952 & 0.762 & 0.948 & 0.854 & 0.726 & 0.528  \\
			ResNet-50 (224)~\cite{ResNet}        & 0.989 & 0.892 & 0.773 &  0.948 & 0.912 & 0.731 & 0.958 & \Red{0.776} & 0.962 & 0.884 & 0.716 & 0.511 \\
			auto-sklearn~\cite{feurer2015efficient}         & 0.934 & 0.716 & 0.649 & 0.779 & 0.902 & 0.719 & 0.887 & 0.601 & 0.942 & 0.855 & 0.690 & 0.515  \\
			AutoKeras~\cite{jin2019auto}           & 0.959 & 0.834 & 0.742 & 0.937 & 0.915 & 0.749 & 0.955 & 0.763 & 0.947 & 0.878 & 0.719 & 0.503 \\
			Google AutoML~\cite{bisong2019google}  & 0.944 & 0.728 &  0.778 &  0.948 & 0.914 &  0.768 & \Blue{0.963} & 0.771 &  0.991 &  0.946 &  0.750 &  0.531 \\
            \rowcolor[HTML]{C8FFFD}
            MedVIT-T (224) & \textcolor{blue}{0.994} & \textcolor{red}{0.938} &  0.786 & \textcolor{red}{0.956} & 0.914 &  0.768 &  \textcolor{red}{0.961} & 0.767 &  \textcolor{red}{0.993} &  \textcolor{red}{0.949} &  0.752 &  0.534 \\
            \rowcolor[HTML]{C8FFFD}
            MedVIT-S (224) & \Red{0.993} & \Blue{0.942} &  \Red{0.791} &  0.954 & \textcolor{blue}{0.937} &  \textcolor{blue}{0.780} &  0.960 & \Blue{0.782} &  \textcolor{blue}{0.995} &  \textcolor{blue}{0.961} &  \Blue{0.773} &  \Blue{0.561} \\
            \rowcolor[HTML]{C8FFFD}
            MedVIT-L (224) & 0.984 & 0.933 &  \textcolor{blue}{0.805} &  \textcolor{blue}{0.959} & \textcolor{red}{0.920} &  \textcolor{red}{0.773} &  0.945 & 0.761 &  0.991 &  0.921 &  \Red{0.754} &  \Red{0.552} \\
			\midrule
			\multirow{2}{*}{\textbf{Methods}} &
			
			\multicolumn{2}{c||}{\textbf{BreastMNIST}} &
			\multicolumn{2}{c||}{\textbf{BloodMNIST}} &
			\multicolumn{2}{c||}{\textbf{TissueMNIST}} &
			\multicolumn{2}{c||}{\textbf{OrganAMNIST}} &
			\multicolumn{2}{c||}{\textbf{OrganCMNIST}} &
			\multicolumn{2}{c}{\textbf{OrganSMNIST}} \\ \cline{2-13}
			 & AUC & ACC & AUC & ACC & AUC & ACC & AUC & ACC & AUC & ACC & AUC & ACC \\ \midrule
			ResNet-18 (28)~\cite{ResNet}          & 0.901 & 0.863 & \Blue{0.998} & 0.958 & 0.930 & 0.676 & \Red{0.997} & 0.935 & 0.992 & 0.900 & 0.972 & 0.782 \\
			ResNet-18 (224)~\cite{ResNet}         & 0.891 & 0.833 & \Blue{0.998} & \Red{0.963} & 0.933 & 0.681 & \Blue{ 0.998} & \Blue{ 0.951} & \Blue{0.994} & \Red{ 0.920} & 0.974 & 0.778 \\
			ResNet-50 (28)~\cite{ResNet}         & 0.857 & 0.812 & \Red{0.997} & 0.956 & 0.931 & 0.680 & \Red{0.997} & 0.935 & 0.992 & 0.905 & 0.972 & 0.770 \\
			ResNet-50 (224)~\cite{ResNet}        & 0.866 & 0.842 & \Red{0.997} & 0.950 & 0.932 & 0.680 & \Blue{ 0.998} & \Red{0.947} & \Red{0.993} & 0.911 & \textcolor{red}{0.975} & 0.785 \\
			auto-sklearn~\cite{feurer2015efficient}        & 0.836 & 0.803 & 0.984 & 0.878 & 0.828 & 0.532 & 0.963 & 0.762 & 0.976 & 0.829 & 0.945 & 0.672 \\
			AutoKeras~\cite{jin2019auto}           & 0.871 & 0.831 & \Blue{0.998} & 0.961 &  0.941 & \textcolor{red}{0.703} & 0.994 & 0.905 & 0.990 & 0.879 & 0.974 & \textcolor{blue}{0.813} \\
			Google AutoML~\cite{bisong2019google} & 0.919 & 0.861 & \Blue{0.998} & \Blue{0.966} & 0.924 & 0.673 & 0.990 & 0.886 & 0.988 & 0.877 & 0.964 & 0.749 \\
            \rowcolor[HTML]{C8FFFD}
            MedVIT-T (224) & \Red{0.934} & \Red{0.896} &  0.996 &  0.950 & \textcolor{red}{0.943} &  \Red{0.703} &  0.995 & 0.931 &  0.991 &  0.901 &  0.972 &  0.789 \\
            \rowcolor[HTML]{C8FFFD}
            MedVIT-S (224) & \Blue{0.938} & \Blue{0.897} &  \Red{0.997} &  0.951& \textcolor{blue}{0.952} &  \textcolor{blue}{0.731} &  0.996 & 0.928 &  \Red{0.993} &  0.916 &  \textcolor{blue}{0.987} &  0.805 \\
            \rowcolor[HTML]{C8FFFD}
            MedVIT-L (224) & 0.929 & 0.883 &  0.996 &  0.954 & 0.935 &  0.699 &  \Red{0.997} & 0.943 &  \Blue{0.994} &  \Blue{0.922} &  0.973 &  \textcolor{red}{0.806} \\
			\bottomrule
		\end{tabular}
	\end{center}
\end{table*}

\subsection{Comparison with State-of-the-art Models}
We compare our MedViT with the latest state-of-the-art methods (e.g. ViTs, CNNs and hybrid networks) with similar model sizes in Table~\ref{tab:Tissue}. We achieve a favorable trade-off between complexity and accuracy. Specifically, our MedViT-T achieves a 70.3\% Top-1 accuracy compared with CNN models, which is better than EfficientNet-B3 and ResNet-18 with more parameters. Similarly, MedViT-S achieves 73.1\% Top-1 accuracy, 5.1\% higher than ResNet-50, 2.6\% higher than EfficientNet-B4, and 0.5\% higher than ConvNext-T, which are famous CNNs. Moreover, MedViT-L outperforms the ConvNext-B by 0.8\%, which has approximately two times more parameters than ours. Furthermore, compared to the pure ViTs, MedViT-T also outperforms PVT-T by a large margin of 6.9\%, while the model complexity is much lower. MedViT-S surpasses Twins-SVT-S by 1\% with similar number parameters. Finally, compared with recent hybrid methods, MedViT-T beats RVT-Ti by 0.7\%. Compared to CvT-13, MedViT-S improve performance by 1.5\% while the complexity is similar. MedViT-L also obtains a 0.6\% performance gain over RVT-B while enjoying less computation complexity. Experimental results illustrate that the proposed MedViT can effectively handle the classification task.

\begin{table}[]
    \centering
    \caption{Classification performance compared with MedViTs and recent state-of-the-art methods on TissueMNIST.}
    \label{tab:Tissue}
    \scalebox{0.75}{
    \begin{tabular}{l|c|cc|c}
    \toprule
    Network  & Image Size   & Param (M)    & FLOPs (G)    &  Top-1 (\%)  \\
    \midrule
    ResNet-18~\cite{ResNet}       & 224    & 11.7      & 1.8       & 68.1          \\
    EfficientNet-B3~\cite{EfficientNet}       & 300    & 12.0      & 1.8      &  69.0          \\
    DeiT-Ti~\cite{Deit}                      & 224       & 5.7    & 1.3       &  59.5                    \\
    PiT-Ti~\cite{PiT}                       & 224       & 4.9    & 0.7       &  62.1                    \\
    PVT-T~\cite{PVT_v1}                        & 224       & 13.2   & 1.9       &  63.4                    \\
    RVT-Ti~\cite{mao2022towards}                       & 224       & 8.6    & 1.3       &  69.6                    \\
    \textbf{MedViT-T} &\textbf{224}    &\textbf{10.8}      &\textbf{1.3}       & \textbf{70.3}        \\
    \midrule

	ResNet-50~\cite{ResNet}       & 224    & 25.6      & 4.1       & 68.0          \\
	EfficientNet-B4~\cite{EfficientNet}       & 380    & 19.3      & 4.2      &  70.5          \\
	ConvNeXt-T~\cite{ConvNext}      &224   & 29.0      & 4.5       &  72.6          \\
    DeiT-S~\cite{Deit}          &224    & 22.0      & 4.6       & 67.0        \\
    Swin-T~\cite{Swin}          &224    & 29.0      & 4.5       & 71.7        \\
    PiT-S~\cite{PiT}                       & 224       & 23.5    & 2.9       &  66.9                    \\
    PVT-S~\cite{PVT_v1}          &224    & 25.4    &4.0  & 66.7        \\
    Twins-SVT-S~\cite{Twins}     &224    & 24.0      & 2.9       & 72.1        \\
    PoolFormer-S36~\cite{metaformer} &224   & 31.2      & 5.0       &  71.8        \\
    CoaT Tiny~\cite{CoaT}  &224 & 5.5      & 4.4       &  69.3        \\
    CvT-13~\cite{CvT}    &224     & 20.1      & 4.5       &  71.6        \\
    RVT-S~\cite{mao2022towards}                       & 224       & 22.1      & 4.7       &  71.2                    \\
    \textbf{MedViT-S} &\textbf{224}    &\textbf{23.6}      &\textbf{4.9}       & \textbf{73.1}        \\

    \midrule
    ResNet-152~\cite{ResNet}       & 224    & 60.2      & 11.3       & 67.5          \\
    ConvNeXt-B~\cite{ConvNext}            &224        & 88.0      & 15.4            & 69.1 \\
    DeiT-B~\cite{Deit}                   &224             & 87.0      & 17.5     & 66.9 \\
    Swin-B~\cite{Swin}              &224  & 87.8      & 15.4       &  68.5        \\
    PiT-B~\cite{PiT}                       & 224       & 73.8    & 12.5       &  68.1                    \\
    PVT-L~\cite{PVT_v1}           &224    &61.4   &9.8  &  66.8        \\
    Twins-SVT-B~\cite{Twins}        &224    & 56.0      & 8.6       &  68.7        \\
    PoolFormer-M36~\cite{metaformer}     &224  & 56.1      & 8.8       & 67.6        \\
    CoaT Small~\cite{CoaT}  &224     & 22.0      & 12.6            & 66.5        \\
    CvT-21~\cite{CvT}       &224     & 32.0      & 7.1       &  67.8        \\
    RVT-B~\cite{mao2022towards}                       & 224       & 86.2      & 17.7       &  69.3                    \\
    \textbf{MedViT-L}     &\textbf{224}  & \textbf{45.8}  & \textbf{13.4} &  \textbf{69.9}        \\
    \bottomrule
    \end{tabular}
    }
\end{table}

\subsection{Average performance }
\label{sec:6.1}
We compare our method with the average AUC and average ACC over all datasets reported in Table~\ref{tab:AVG_2DResults}. Our models shows average AUCs 93.6\%, 94.2\%, 93.5\% and average ACCs 84\%, 85.1\%, 84.2\% in the total of twelve different datasets obtained by MedViT-T, MedViT-S and MedViT-L, respectively. MedViT-S outperforms all the baseline ResNets and AutoML methods in both average AUC and average ACC by a large margin, demonstrating the advantage of using transformer vision to classify medical image.

In Table~\ref{tab:AVG_2DResults}, we also compare the numbers of parameters of our proposed method with baseline ResNets. Our MedViT shows great superiority in terms of performance while the model complexity of ours is on par with baseline ResNets.

\begin{table}[h]
	\centering
	\small
	\tablestyle{5pt}{1.05}
	\caption{ Average performance comparison in standard metrics of average ACC and average AUC over all MedMNIST-2D.}
	\label{tab:AVG_2DResults}
		\begin{tabular}{l|c|cc}
			\toprule
			\multirow{2}{*}{Methods} & \textbf{Params} & \multicolumn{2}{c}{\textbf{Avg.}} \\
                                     & (M)             & AUC        & ACC                  \\
			\midrule
			ResNet-18 (28)~\cite{ResNet}     & 11.2 &0.922 & 0.819  \\
			ResNet-18 (224)~\cite{ResNet}        & 11.2 & 0.925 &  0.821  \\
			ResNet-50 (28)~\cite{ResNet}         & 23.5 & 0.920 & 0.816  \\
			ResNet-50 (224)~\cite{ResNet}        & 23.5 & 0.923 &  0.821  \\
			auto-sklearn~\cite{feurer2015efficient}        & - & 0.878 & 0.722  \\
			AutoKeras~\cite{jin2019auto}           & - & 0.917 & 0.813  \\
			Google AutoML~\cite{bisong2019google}  & - &  0.927 & 0.809 \\
            \midrule
            MedViT-T (224) & 10.2 &  0.936 & 0.840 \\
            MedViT-S (224) & \textbf{23} &  \textbf{0.942} & \textbf{0.851} \\
            MedViT-L (224) & 45 &  0.935 & 0.842 \\
			\bottomrule
		\end{tabular}
\vspace{-10pt}
\end{table}

\subsection{Visual inspection of MedViT}
To further verify the property of our MedViT, we apply Grad-CAM~\cite{selvaraju2017grad} on the ESA’s output in the last ECB to qualitatively inspect MedViT. We visualize the heat maps of the output features from ResNet-18 and MedViT-T in Figure~\ref{fig:heatmap}. Compared with the baseline ResNet-18, our MedViT covers the relevant locations in the images more precisely and attends less to the background. Moreover, MedViT can better handle the scale variance issue as shown in Derma, Oct and Path. That is, it covers target area accurately whether they are small, medium, or large in size. In the Retina dataset, it can be seen that our model in the heat map of retinal fundus image can well recognize the direction and area of the specific lesion. Our model well-localized a focal infected area by bacterial infection in the heat map of Chest dataset, while it was also able to delineate the multi-focal lesions in periphery of both upper lungs in the heat map of Pneumonia dataset, which is typical findings for pneumonia.

Such observations demonstrate that introducing the intrinsic IBs of locality and scale-invariance from convolutions to transformers helps MedViT learn capable of simultaneously capturing high-quality and multi- frequency signals. Compared to the conventional approach, our method mitigates the background bias significantly.

\subsection{Augmentation and Robustness Evaluation}
To evaluate our model against the adversarial attack benchmarks, we adopt a common gradient-based attack  method FGSM~\cite{goodfellow2014explaining} and a powerful multi-step attack PGD~\cite{madry2017towards} with a step size of $4/255 = 0.015$ and steps $n_{iter} = 5$. For both attackers, the magnitude of the adversarial noise is $\varepsilon$ of $8/255 = 0.031$. Results in Table~\ref{tab:adversarial comparison} demonstrate that different blocks of MedViT architecture have a strong correlation with the adversarial robustness. The proposed MedViT-T and MedViT-T* represent high adversarial robustness under both attack benchmarks. This is ascribed to the Efficient Convolution Block and Patch Momentum Changer, which aims to improve the robustness of medical diagnostic. In ECB, adding depth-wise convolution into feed-forward networks help model to better capture local dependencies within tokens. Moreover, the PMC module utilizes implicit data augmentation at token-level, which forces ViTs to pay special attention towards local features at different stages. We show empirically that MedViT by using these blocks is consistently able to improve robustness and classification accuracy across medical datasets.

The proposed MedViT-T* model achieves superior performance on both admired PGD and FGSM attacks in compared ResNet and baseline MedViT. In detail, MedViT-T* considerably outperforms the counterparts in TissueMNIST with gains of 38.4\%, 6.1\% on FGSM attack and gains of 30.2\%, 7.5\% on PGD attack compared to ResNet-18 and MedViT-T, respectively. MedViT-T* also achieves outstanding standard performance (ACC) consistently on four MNIST dataset. Specifically, the Transformer Augmentation Block module brings significant improvements (1.5\%, 3.1\%, 1.1\% and 0.7\%) on the Oct, Tissue, Retina and Path MNIST datasets, respectively. This advance is further expanded by our Locally-FeedForward and PMC augmentation. Nonetheless, our MedViT-T$^*$ model generally yields the best accuracy/robustness tradeoff.

\begin{table*}[!ht]
    \centering
    \caption{The performance of MedViT-T, MedViT-T$^*$ and ResNet-18 on four MedMNIST-2D and two robustness benchmarks. Except for MedViT-T$^*$ architecture, we do not make use of any specialized modules or additional fine-tuning procedures.}
    \label{tab:adversarial comparison}
    \resizebox{\textwidth}{!}{
    \begin{tabular}{l||ccc||ccc||ccc||ccc}
    \toprule
    \multirow{2}{*}{\textbf{Methods}} & \multicolumn{3}{c||}{\textbf{OctMNIST}} &  \multicolumn{3}{c||}{\textbf{TissueMNIST}}  & \multicolumn{3}{c||}{\textbf{RetinaMNIST}} & \multicolumn{3}{c}{\textbf{PathMNIST}} \\
    \cline{2-13}
     & \textbf{ACC} & \textbf{FGSM} & \textbf{PGD} & \textbf{ACC} & \textbf{FGSM} & \textbf{PGD} & \textbf{ACC} & \textbf{FGSM} & \textbf{PGD} & \textbf{ACC} & \textbf{FGSM} & \textbf{PGD} \\
    \midrule
    ResNet-18 (224) & 0.763 & 0.238 & 0.201 & 0.681 & 0.096 & 0.090 & 0.493 & 0.167 & 0.117 & 0.909  & 0.406  & 0.108
    \\
    MedViT-T (224) & 0.767 & 0.272 & 0.249 & 0.703 & 0.419 & 0.317 & 0.534 & 0.168 & 0.145 & 0.938  & 0.562  & 0.224
    \\
    MedViT-T$^*$ (224) & \textbf{0.782} & \textbf{0.304} &  \textbf{0.297} &  \textbf{0.734} &  \textbf{0.480} &  \textbf{0.392} &  \textbf{0.545} & \textbf{0.197} &  \textbf{0.180} &  \textbf{0.945}  &  \textbf{0.585}  &  \textbf{0.245}
    \\
    \bottomrule
    \end{tabular}
    }
    \vspace{-1em}
\end{table*}
\vspace{-1em}

\subsection{Ablation Study}
We conduct various ablation experiments to investigate the effectiveness of the critical blocks of our architecture. Firstly, we study the impact of Efficient Convolution Block on robust and clean accuracy in comparison with the most well-known components. Afterwards, we individually evaluate Patch Moment Changer block at different stages of our architecture. It is important to note that all of our ablation experiments are based on the MedViT-T model on TissueMNIST.

\textbf{Impact of Efficient Convolution Block.}
To analyze the effectiveness of our ECB for improving robustness/accuracy of Transformers, we substitute ECB in MedViT with famous blocks, including ConvNext~\cite{ConvNext} block, Residual block in ResNet~\cite{ResNet}, PoolFormer Block~\cite{metaformer}, and LSA block in Twins~\cite{Twins}. We constantly keep other components of our architecture unchanged to build different models under similar complexity. As illustrated in Table~\ref{table:7}, our architecture with ECB block achieves the best robustness/accuracy in comparison with prior blocks. In particular, ECB outperforms the ConvNext block (runner-up) by 0.6\% in clean and 4.4\% in robust accuracy with lower model complexity.
\begin{table}[h!]
	\vspace{5pt}
    \small
    \centering
    \caption{\textbf{Impact of Efficient Convolution Block.} Performance of clean accuracy and adversarial robustness under FGSM attack on TissueMNIST.}
     \tablestyle{5pt}{1.05}
\begin{tabular}{c|cc|c|c}
\toprule

\multirow{2}{*}{Block type} & \multicolumn{2}{c|}{Model Complexity} & Clean & Robust \\
                            & Params (M) & Flops (G) & Acc(\%) & Acc(\%)   \\
\midrule
Residual Block~\cite{ResNet} & 10.9 & 1.3 & 68.1 & 22.3 \\
ConvNeXt Block~\cite{ConvNext} & 11.2 & 1.4 & 69.7 & 37.5 \\
PoolFormer Block~\cite{metaformer} & 10.7 & 1.1 & 68.9 & 29.1 \\
LSA Block~\cite{Twins} & 12.7 & 2.1 & 69.2  & 31.8 \\
\textbf{ECB (ours)} & \textbf{10.8} & \textbf{1.3} & \textbf{70.3} & \textbf{41.9} \\

\bottomrule
\end{tabular}
\label{table:7}
\vspace{5pt}
\end{table}

\textbf{Effect of PMC in Different Stages.}
To find the best place for Patch Moment Changer in our model, we combine the efficient blocks with PMC in different stages. PMC is applied to 3 different stages of MedViT-T on TissueMNIST to find the best setup.
As illustrated in Table~\ref{table:8}, PMC works best when applied after the first stage of the 4-stage MedViT-T. We assume PMC helps transformers to capture local information and has a significant advantage at the early stages. In contrast, the last stages already contain a lot of information, which is less impacted by the effect of PMC. In this paper, we adopt PMC augmentation in the first stage.

\begin{table}[h!]
	\vspace{5pt}
    \small
    \centering
    \caption{\textbf{Effect of PMC in Different Stages.} Performance (\%) of clean accuracy and adversarial robustness under FGSM attack on TissueMNIST. Place of PMC is indicated by \cmark in different stages.}
     \tablestyle{5pt}{1.05}
\begin{tabular}{c|c|c|c|c}
\toprule

\multicolumn{3}{c|}{Augmentations} & \multirow{2}{*}{Acc} & \multirow{2}{*}{Rob. Acc} \\
 Stage 1 & Stage 2 & Stage 3 & & \\
\midrule
\xmark & \xmark & \xmark  & 70.3 & 41.9 \\
\cmark & \xmark & \xmark & \textbf{73.4} & \textbf{48.0} \\
\xmark & \cmark & \xmark & 72.9 & 45.4 \\
\xmark & \xmark & \cmark & 71.5  & 44.1 \\
\bottomrule
\end{tabular}
\label{table:8}
\vspace{5pt}
\end{table}

\section{Conclusion}

In this paper, we introduce a family of MedViT, a novel hybrid CNN-transformer architecture for medical image classification. Specifically, we combine the local representations and the global features by using robust components. Furthermore, we have devised a novel patch moment changer augmentation that adds rich diversity and affinity to training data. Experiments show that our MedViT achieves state-of-the-art accuracy and robustness on the standard large-scale collection of 2D biomedical datasets. We hope that our model can encourage more researchers and provide inspire for future works on realistic medical deployment.

\printcredits

%
\bibliographystyle{IEEEtran} 				
\bibliography{cas-refs}



\end{document}